\documentclass[conference]{IEEEtran}
\IEEEoverridecommandlockouts
\usepackage{cite}
\usepackage{amsmath,amssymb,amsfonts}
\usepackage{graphicx}
\usepackage{textcomp}
\usepackage{algorithm}
\usepackage{algpseudocode}
\usepackage{graphicx} 
\usepackage{listings}
\usepackage{tcolorbox}
\usepackage{amsmath}
\usepackage{amsfonts}

\usepackage{algpseudocode}
\usepackage{amsmath}
\usepackage{booktabs}
\usepackage{multirow}
\usepackage{subcaption}
\usepackage{array}
\usepackage{booktabs}
\usepackage{fancyhdr}
\usepackage{url}

\def\BibTeX{{\rm B\kern-.05em{\sc i\kern-.025em b}\kern-.08em
    T\kern-.1667em\lower.7ex\hbox{E}\kern-.125emX}}
\begin{document}

\title{Accelerating Complex Disease Treatment through Network Medicine and GenAI: A Case Study on Drug Repurposing for Breast Cancer}

\author{\IEEEauthorblockN{1\textsuperscript{st} Ahmed Abdeen Hamed}
\IEEEauthorblockA{\textit{Dept. of SSIE} \\
\textit{Binghamton University}\\
% Binghamton, United States \\
and \\
iMol Polish Academy of Sciences\\
Warsaw, Poland\\
ahamed1@binghamton.edu*}
\and
\IEEEauthorblockN{2\textsuperscript{nd} Tamer E. Fandy}
\IEEEauthorblockA{\textit{Dept. of Medical Education} \\
\textit{Texas Tech University HSC}\\
El Paso, United States \\
Tamer.fandy@ttuhsc.edu}
% \and
% \IEEEauthorblockN{3\textsuperscript{rd} Xindong Wu}
% \IEEEauthorblockA{\textit{Dept. Key Lab of KEBD} \\
% \textit{Hefei University of Technology}\\
% Hefei, China \\
% xwu@hfut.edu.cn}
}

\maketitle

\begin{abstract}
The objective of this research is to introduce a network specialized in predicting drugs that can be repurposed by investigating real-world evidence sources, such as clinical trials and biomedical literature. Specifically, it aims to generate drug combination therapies for complex diseases (e.g., cancer, Alzheimer's). We present a multilayered network medicine approach, empowered by a highly configured ChatGPT prompt engineering system, which is constructed on the fly to extract drug mentions in clinical trials. Additionally, we introduce a novel algorithm that connects real-world evidence with disease-specific signaling pathways (e.g., KEGG database). This sheds light on the repurposability of drugs if they are found to bind with one or more protein constituents of a signaling pathway. To demonstrate, we instantiated the framework for breast cancer and found that, out of 46 breast cancer signaling pathways, the framework identified 38 pathways that were covered by at least two drugs. This evidence signals the potential for combining those drugs. Specifically, the most covered signaling pathway, ID hsa:2064, was covered by 108 drugs, some of which can be combined. Conversely, the signaling pathway ID hsa:1499 was covered by only two drugs, indicating a significant gap for further research. Our network medicine framework, empowered by GenAI, shows promise in identifying drug combinations with a high degree of specificity, knowing the exact signaling pathways and proteins that serve as targets. It is noteworthy that ChatGPT successfully accelerated the process of identifying drug mentions in clinical trials, though further investigations are required to determine the relationships among the drug mentions. 
\end{abstract}


\section{Introduction}

\subsection{Highlights}
\begin{enumerate}
    \item[1. ] Presenting a multilayered network medicine approach for complex diseases, empowered by Generative AI, that produces drug repurposing combinations from real-world evidence.
    \item[2. ] Introducing a highly configurable learning algorithm that harnesses ChatGPT prompt engineering, executed on the fly to analyze each data item (clinical trials) with the help of a few-shot examples.
    \item[3. ] Demonstrating the full potential of the algorithm in a breast cancer case study, showing how it produces combinations with a high degree of specificity in terms of drug targets.
\end{enumerate}

\subsection{Literature Review}
Network Medicine, a well-established research area, has made significant contributions to understanding various human diseases, biology, drug treatments, and biological targets~\cite{maron2020global, barabasi2011network}. In recent years, such networks have proven to be reliable platforms for drug repurposing, especially for complex diseases~\cite{sadegh2021network}. This was particularly evident during the Coronavirus pandemic, when various research efforts leveraged network medicine to develop viable treatments in response to the global crisis~\cite{sadegh2020exploring, hamed2022covid, hamed2022mining, morselli2021network, gates2020anatomy, zhou2020network, de2022machine}.

With the emergence of Generative AI (GenAI) and Large Language Models (LLMs), new opportunities driven by ChatGPT~\cite{chatgpt} are on the rise~\cite{xu2024chatgpt, rodriguez2024leveraging, savage2023drug, chen2024multi, callaway2024chatgpt, sharma2024assessment}. In this work, we focus on multilayered network medicine and present some of the latest and most relevant research related to our approach.

Rocha et al. presented a multilayered personalized health platform that integrates various heterogeneous resources to provide insights for epilepsy management~\cite{correia2024myaura}. Their approach integrated data from electronic health records, social media platforms, and biomedical databases.

Kim et al. developed a multi-layered knowledge graph and employed a neural network approach to predict the recurrence risk of a hormone receptor related to breast cancer~\cite{lee2024multi}.

Xie et al. reviewed how heterogeneous multilayered knowledge graphs are used to integrate and analyze omics data produced from gene sequencing and high-throughput techniques~\cite{lee2020heterogeneous}.

McLean reviewed the use of multi-modal knowledge graphs as tools for drug discovery, particularly in the context of COVID-19~\cite{maclean2021knowledge}.

Kim et al. presented a semantic multilayered knowledge graph for drug repurposing. They utilized the "guilt-by-association" principle to recommend drug candidates for given diseases in the network, employing semantically guided random walks to achieve the desired results~\cite{bang2023biomedical}.

While multilayered knowledge graphs are powerful, integrating heterogeneous datasets and constructing linked graph layers are complex and challenging processes. In this paper, we aim to accelerate these processes to enable our network medicine platform to address drug repurposing questions, specifically demonstrating capabilities for breast cancer. We also explore how ChatGPT, as a GenAI tool, was used via prompt engineering to perform similar tasks.
\\
Polak and Dane introduced a tool called ChatExtract, which performs information extraction from research papers using a set of engineered prompts. These prompts guide conversations, extract data, and enforce correctness through follow-up questions. The authors claimed that their approach ensured outstanding performance and accurate results~\cite{polak2024extracting}.
\\
Wang et al. examined the reliability of prompt engineering using different styles for five tasks related to the American Academy of Orthopedic Surgeons (AAOS) osteoarthritis (OA) evidence-based guidelines. Their study revealed that carefully engineered prompts could improve the accuracy of responses to professional medical questions and the information extracted~\cite{wang2024prompt}.
\\
Snyder et al. applied LLM prompt engineering using a few-shot learning technique for various drug discovery applications. They compared this approach with classical machine learning techniques and concluded that while classical methods are best for large and diverse datasets, LLMs can provide comparable results when analyzing small and homogeneous datasets~\cite{snyder2024goldilocks}.

With this introduction, we propose a multilayered network medicine approach, accelerated by ChatGPT few-shot prompt engineering, that streamlines data processing and network layer construction without compromising the accuracy of results.

\section{Data}
This research used three different type of resources: 

\begin{itemize}
    \item[1. ] \textbf{Clinical Trials as a Source of Real-World Evidence:} We searched for drug combinations using the query "drug combination" and acquired approximately 2,450 trials that mentioned treatments involving more than one drug. The trials were extracted in JSON format, with each trial stored in an independent file.
    \item[2. ] \textbf{KEGG Breast Cancer Signaling Pathways as Potential Targets:} The signaling pathways serve two purposes: first, providing a list of relevant biological targets; second, identifying drugs that may bind with proteins in the same signaling pathway to offer combination insights.
    \item[3. ] \textbf{BioMedical Publications Retrieved Using PubMed APIs:} To identify drug-target connections, we used the eSearch command-line utility to execute the query "breast cancer." This resulted in acquiring 440,000 MEDLINE records containing the PubMed ID, title, and abstract.
\end{itemize}
\section{Methods}
The challenge of identifying drug combination opportunities for complex diseases in general and breast cancer in particular requires harvesting knowledge from multiple sources. This calls for a multifaceted framework, centered around layered network medicine, to address various tasks:
\begin{enumerate}
    \item[1.] \textbf{Drug Combos Layer:} Using manually annotated clinical trial descriptions, we engineered a ChatGPT model with few-shot learning to recognize drugs and identify keywords and symbols that indicate the combination of two or more drugs in a trial. Examples such as ``combined with,'' ``in addition to,'' and "+" were included in the training examples. The results were pruned using the drug branch in the ChEBI ontology \cite{hull2010chebi}.
    \item[2.] \textbf{Drug-Target Layer:} The drug combinations identified in the previous layer serve as the source elements of the graph, while the targets are elements in the breast cancer KEGG signaling pathways. The links between these entities are captured based on the proximity distance in biomedical literature abstracts. Proximity is also commonly used in feature extraction from text \cite{hamed2023detection}, which is foundational in network medicine \cite{gan2023network, correia2024myaura, do2021network}. 
    \item[3.] \textbf{Biological Target Signaling Pathways} The success of this work hinges on recommending viable drug combinations. Therefore, we introduce an algorithmic approach that explores the various layers and produces recommendations with a high degree of specificity, and may be ready to be tested using in-vitro and clinical trials.
\end{enumerate}

\subsection{Drug Combinations Layer}
The description section of clinical trials is a rich source of knowledge and presents various clinical opportunities. By mining these resources, we can gain useful insights (e.g., the medical condition, treatment, indication, protocol followed, etc). To start, we extracted the description field from the JSON representation of the NCT records. We then hand-picked a few representative examples from trials that contained various drugs for different conditions, where the trials also investigated potential drug combinations. Each selected example signaled the use of combinations in a unique way. For instance, while one trial explicitly mentioned a drug to be combined using expressions like combined with,'' co-administered with,'' and ``in addition to,'' other trials used symbols such as ``+'' and ``PLUS.''

To enable ChatGPT to recognize the drugs and combination signaling keywords, we manually annotated seven distinct examples as few-shot learning prompts for the prompt engineering algorithm. Specifically, we used different notations to distinguish drug mentions from combination signaling keywords. Drug mentions were annotated using the angle brackets ``$\langle$'' and ``$\rangle$'', while combination keywords were annotated using square brackets ``['' and ``]''. Below are some concrete examples we used to train ChatGPT, illustrating the annotation method (refer to \ref{box:examples}2). 

\begin{tcolorbox}[colback=blue!5!white, colframe=blue!75!black, title=Selected Samples of Training Examples, label=box:examples]
    \begin{itemize}
        \item \textcolor{orange}{$\langle$oseltamivir$\rangle$} \textcolor{blue}{[+]} \textcolor{orange}{$\langle$zanamivir$\rangle$}
        \item versus \textcolor{orange}{$\langle$docetaxel$\rangle$} in combination with either \textcolor{orange}{$\langle$gemcitabine$\rangle$} or \textcolor{orange}{$\langle$vinorelbine$\rangle$} or in \textcolor{blue}{[combination]} with \textcolor{orange}{$\langle$capecitabine$\rangle$}.
        \item \textcolor{orange}{$\langle$zidovudine$\rangle$} ( AZT ) \textcolor{blue}{[plus]} \textcolor{orange}{$\langle$didanosine$\rangle$} (ddI), AZT plus \textcolor{orange}{$\langle$zalcitabine$\rangle$} (ddC), AZT alternating monthly with ddI, and AZT/ddI \textcolor{blue}{[plus]} \textcolor{orange}{$\langle$nevirapine$\rangle$}
        \item \textcolor{orange}{$\langle$SOT102$\rangle$} administered as monotherapy (Part A) and in \textcolor{blue}{[combination]} with first-line SoC treatment \textcolor{orange}{$\langle$mFOLFOX6$\rangle$} with \textcolor{orange}{$\langle$nivolumab$\rangle$} and nab-paclitaxel/ \textcolor{orange}{$\langle$gemcitabine$\rangle$};
        \item \textcolor{orange}{$\langle$GW642444M$\rangle$} and \textcolor{orange}{$\langle$GW685698X$\rangle$} will be simultaneously \textcolor{blue}{[co-administered]} from a single device and compared with GW642444M and GW685698X administered separately in order to determine whether co-administration affects the safety
    \end{itemize}
\end{tcolorbox}

As for the actual prompt itself, it contained three types of information: (1) a system role, where the content instructed ChatGPT to adopt the role of a subject matter expert capable of recognizing drugs and combinations from clinical trials; (2) a user role, where the user instructs ChatGPT to identify drugs and combinations from unseen clinical trial descriptions by learning from the seven manually annotated examples provided; (3) an assistant role, which provides new but unannotated clinical trial descriptions. The prompt required the return of drugs and their combinations from each trial when found. It also instructed the ChatGPT engine that each drug combination to be written in a single line for easier post-processing. This was demonstrated to ChatGPT using two examples which we also injected as part of the prompt. Therefore, the prompt engineering algorithm expects various user inputs to be used in order: (a) the list of annotated examples used as few-shots for ChatGPT to learn from, (b) one clinical trial description at a time to be analyzed, (c) the ID of the clinical trial being processed, and (d) training output examples for ChatGPT to produce the output accordingly. These steps are described in detail in Algorithm \ref{alg:drug_combination}.

\begin{algorithm}
\caption{Drug and Combination Detection}
\label{alg:drug_combination}
\begin{algorithmic}[1]
\State \textbf{Input:} $nct\_dataset$ dataset
\State \textbf{Input:} [\texttt{shot\_0}, \dots, \texttt{shot\_6}] example annotations
\State \textbf{Input:} [\texttt{comb\_ex1}, \texttt{comb\_ex2}] example output
\State \textbf{Output:} \textit{response} (ChatGPT response with detected drug combinations)

\For{\textbf{each} $(nct\_desc)$ \textbf{in} $(nct\_dataset)$ }
    \State \textbf{prompt} $\gets$ 
    \Statex \hspace{2em} \texttt{"""}
    \Statex \texttt{You are a specialized drug annotator, detect drugs if annotated/marked using < and > and relationship if it is annotated/marked using [ and ] 
    \Statex
    \Statex Your task is to learn from the following few shots:}
    \Statex \hspace{2em} \texttt{\{shot\_0\}}
    \Statex \hspace{2em} \texttt{\{shot\_1\}}
    \Statex \hspace{2em} \texttt{\dots}
    \Statex \hspace{2em} \texttt{\{shot\_6\}}
    \Statex
    \Statex \texttt{Now you need to analyze the description of this clinical trial nct\_desc to identify drugs and potential combinations}
    \Statex
    \Statex \texttt{Your response will be combinations discovered in pipe-delimited format that follows the following examples:}
    \Statex \hspace{2em} \texttt{\{nct\_id | comb\_ex1\}}             
    \Statex \hspace{2em} \texttt{\{nct\_id | comb\_ex2\}} 
    \Statex
    \Statex \texttt{Write each combination found in a single line and no other messages should be written}
    \Statex \hspace{2em} \texttt{"""}
    \Statex
    \State \textit{response} $\gets$ \textbf{client.chat.completions.create}(
    \Statex \hspace{4em} \textbf{model} = "gpt-3.5-turbo",
    \Statex \hspace{4em} \textbf{messages} = [
    \Statex \hspace{6em} \{\textbf{"role"}: \textbf{"user"}, \textbf{"content"}: \textit{prompt}\}
    \Statex \hspace{4em} ]
    \Statex )
\EndFor
\end{algorithmic}
\end{algorithm}

With the ChatGPT prompt engineering task is performed, and the output is captured, we have a pipe-delimited dataset of triples \(\langle \text{$NCT_n$}, \text{$d_i$}, \text{$d_j$} \rangle\), where the first element is the clinical trial ID, and the next two elements are the two members of a combination in a clinical trial $n$. The process of constructing the drug combination layer is as follows:

Let \( \mathcal{N} \) be the set of clinical trials.
Each clinical trial \( n \in \mathcal{N} \) has a set of drug mentions \( D_n = \{ d_1, d_2, \ldots, d_{x_n} \} \), where \( d_i \) represents a drug.
The combinations of these drugs form edges in the graph.
Each edge is labeled with the clinical trial ID from which it originated.

\begin{itemize}
    \item \textbf{Clinical Trials:} 
    \[
    \mathcal{N} = \{ n_0, n_2, \ldots, n_{N-1} \}
    \]
    where \( N \) is the total number of clinical trials.
    
    \item \textbf{Drug Mentions:} For each clinical trial \( n \in \mathcal{N} \),
    \[
    D_n = \{ d_0^n, d_1^n, \ldots, d_{x-1}^n \}
    \]
    where \( D_n \) is the set of drug mentions in clinical trial \( n \), and \( {x-1}_n \) is the number of drug mentions in \( n \).
    
    \item \textbf{Drug Combinations:} The combinations of drugs in clinical trial \( n \) form pairs \( (d_i^n, d_j^n) \) with \( i \neq j \). These combinations can be represented as a set of edges \( E_n \):
    \[
    E_n = \{ (d_i^n, d_j^n) \mid d_i^n, d_j^n \in D_n, i \neq j \}
    \]
    
    \item \textbf{Graph Representation:} The entire set of drugs \( D \) across all clinical trials forms the nodes of the graph:
    \[
    D = \bigcup_{n \in \mathcal{N}} D_n
    \]
    The edges \( E \) are formed from the combinations in each clinical trial and are labeled with the trial ID:
    \[
    E = \bigcup_{n \in \mathcal{N}} \{ ((d_i^n, d_j^n), n) \mid (d_i^n, d_j^n) \in E_n \}
    \]
    
    \item \textbf{Multi-Edge Graph:} Let \( G = (V, E) \) be a multi-edge graph where \( V \) is the set of drug nodes and \( E \) is the set of edges labeled with the clinical trial IDs:
    \[
    V = D
    \]
    \[
    E = \{ ((d_i, d_j), n) \mid \exists n \in \mathcal{N} \text{ such that } (d_i, d_j) \in E_n \}
    \]
    where \( ((d_i, d_j), n) \) indicates an edge between drug \( d_i \) and drug \( d_j \) labeled with the clinical trial ID \( n \).
\end{itemize}

\subsection{Drug-Target Layer}
\begin{figure}[h]
  \centering
  \includegraphics[scale=0.15]{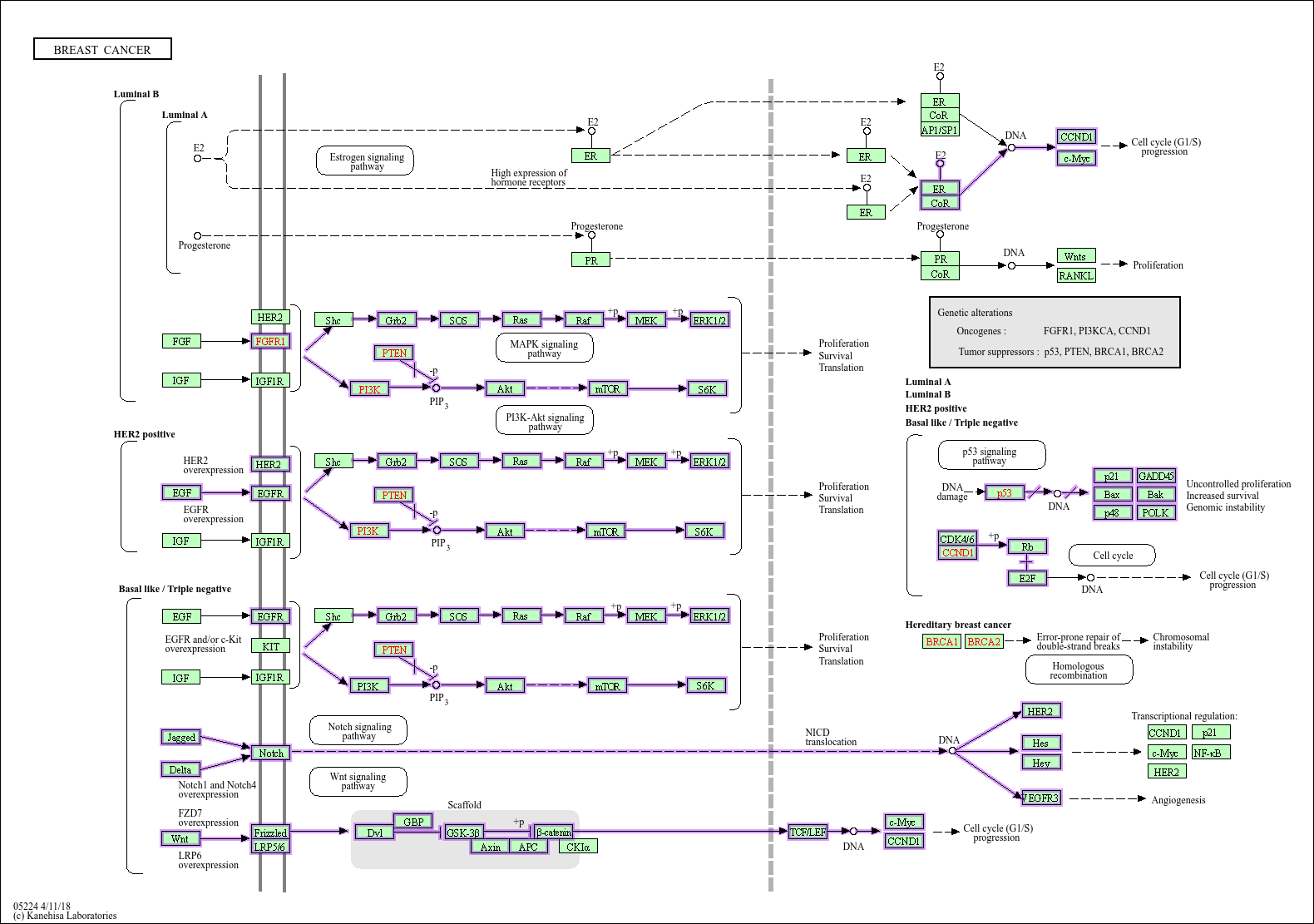}
  \caption{The KEGG Pathways for the Breast Cancer Module}\label{fig:bcpathways}
\end{figure}

The description sections of clinical trials are rich with drug names and their combinations. However, they often lack information about drug targets, which is a key indication for a drug to be repurposed. Drug-target interaction is essential in drug repurposing \cite{ye2021unified}. Therefore, it is crucial to explore the signaling pathways between drugs and targets to predict such repurposeable targets \cite{gkr912_stuperTarget}. Since biomedical literature is rich in such knowledge \cite{hong2020novel}, we employ a text-mining approach to identify drug names, drug targets, and possible interactions. Specifically, we use the Chemical Entities of Biological Interest (ChEBI) to identify drug names \cite{hull2010chebi} and the proteins that constitute a cancer-type module in the KEGG signaling pathways database (e.g., breast cancer) to identify protein mentions in the abstracts \cite{kanehisa2000kegg}. While the cancer-disease pathways play a significant role in predicting repurposable combinations, we utilize them individually as independent entities. The following tasks are involved in constructing this layer:
\begin{enumerate}
    \item[1.] \textbf{Abstract Text Processing:} This includes sentence and word tokenization, and case normalization. For this task, we used the \textit{word\_tokenize} and \textit{sent\_tokenize} modules from the ``nltk.tokenize'' Python package.
    
    \item[2.] \textbf{ChEBI Ontology for Drug Search:} The identification of drugs requires a dictionary look-up mechanism where PubMed abstract words must be an identical match with one of the given terms.

    \item[3.] \textbf{Cancer Diseases KEGG Signaling Pathways Pre-Processing:} The breast cancer pathways branches are XML resources that require parsing. Starting from the root, we traverse the XML tree to search for elements of type ``Gene.'' We extract two bits of information: (1) the ``name'' of the branch, which has a unique ID, and (2) a set of proteins, known as ``graphics,'' which make up each pathway as part of the cancer module. The total number of unique targets found associated with breast cancer is 383 proteins. Figure \ref{fig:bcpathways} shows the various signaling pathways as part of the Breast Cancer KEGG module 
    
    \item[4.] \textbf{Keeping Track of Drug-Target Proximity:} This task requires measuring the proximity of a drug to proteins that are part of the pathways extracted in the step above. By capturing the index of each term, we can measure the distance in token units.
    
    \item[5.] \textbf{Constructing the Drug-Target Layer:} Using the proximity as a condition to satisfy, we capture drug-target pairs that are found within the given proximity. They contribute a node of their own type (drug and protein) and a link of type proximity to the network layer. The network construction task is formally described in Algorithm \ref{alg:drug_target_graph}, and is implemented using the using the ``networkx'' python package \cite{osti_960616}. 
\end{enumerate}.

\begin{algorithm}
\caption{Generate Graph of Drug-Target Pairs from PubMed Dataset}
\label{alg:drug_target_graph}
\begin{algorithmic}[1]
\State \textbf{Input:} $\mathcal{C}$: CHEBI drug list
\State \textbf{Input:} $\mathcal{P}$: List of breast cancer proteins
\State \textbf{Output:} $\mathcal{G}$: Graph of drug-target pairs

\State Initialize $\mathcal{G} \gets (\mathcal{V}, \mathcal{E})$
\Comment{Graph with nodes $\mathcal{V}$ and edges $\mathcal{E}$}
\State $\mathcal{V} \gets \emptyset$
\Comment{Initialize set of vertices}
\State $\mathcal{E} \gets \emptyset$
\Comment{Initialize set of edges}

\For{each $(\delta, \alpha)$ in PubMed dataset}
    \State $\tau \gets \text{tokenize}(\alpha)$
    \Comment{Tokenize abstract $\alpha$}
    \State $\text{doc\_edges} \gets []$
    \Comment{Initialize document edge list}
    \For{each $d \in \mathcal{C}$}
        \For{each $p \in \mathcal{P}$}
            \If{$d \in \tau$ \textbf{and} $p \in \tau$}
                \State $\pi \gets \text{calc\_distance}(d, p)$
                \Comment{Calculate proximity $\pi$}
                \State $\epsilon \gets (d, p)$
                \Comment{Create drug-protein link}
                \State $\lambda \gets (\delta, \epsilon, \pi)$
                \Comment{Create a proximity edge}
                \State $\text{doc\_edges} \gets \text{doc\_edges} \cup \{\lambda\}$
                \State $\mathcal{V} \gets \mathcal{V} \cup \{d, p\}$
                \Comment{Add $d$ and $p$ to vertices}
                \State $\mathcal{E} \gets \mathcal{E} \cup \{\epsilon\}$
                \Comment{Add $\epsilon$ to edges}
            \EndIf
        \EndFor
    \EndFor
\EndFor
\State $\mathcal{G} \gets (\mathcal{V}, \mathcal{E})$
\Comment{Finalize graph $\mathcal{G}$}
\State \textbf{return} $\mathcal{G}$
\end{algorithmic}
\end{algorithm}

\subsection{Biological Targets Pathways Layer} 
Upon successfully constructing a drug combination layer from clinical trials and a drug-target layer from biomedical literature, we require additional information to determine if a drug-target link is cancer-related. This crucial information can be derived from the breast cancer KEGG pathways, conditioned on identifying a target present in the drug-target layer that also corresponds to one of the cancer biological pathways extracted from the KEGG database. The algorithmic process consists of three tasks:

\begin{enumerate}
    \item[1.] Starting with the drug-target layer, examine the target nodes of each edge.
    
    \item[2.] Verify whether the target exists in a cancer pathway. If two or more different drugs are found connected with different proteins of the the same pathway, store them as potential candidates.
    
    \item[3.] Search the drug combination layer to determine if the drug candidates identified in step 2 also form a combination in clinical trials. If they do, store them as evidence for repurposing; otherwise, store them as potential combinations pending further investigation.

\end{enumerate}

\begin{algorithm}
\caption{Generate Drug Combinations Evidence Graph}
\label{alg:drug_combinations_graph}
\begin{algorithmic}[1]
\State \textbf{Input:} $\mathcal{L}_1$: Drug-target network layer
\State \textbf{Input:} $\mathcal{L}_2$: Drug combination network layer
\State \textbf{Input:} $\mathcal{C}$: Cancer disease pathway layer
\State \textbf{Output:} $\mathcal{R}$: Drug combinations evidence graph
\State Initialize $\Gamma \gets \emptyset$
\Comment{Temporary graph storage}
\State Initialize $\mathcal{R} \gets \emptyset$
\Comment{Result graph}
\ForAll{$e \in \mathcal{L}_1$}
    \ForAll{$c \in \mathcal{L}_2$}
        \ForAll{$d \in e$}
            \If{$d \in \mathcal{L}_2$}
                \State $\Gamma \gets \Gamma \cup \{(d, \tau, \nu)\}$
                \Comment{$\nu$: Clinical trial ID, $\tau$: Target}
            \Else
                \State $\Gamma \gets \Gamma \cup \{(d, \tau)\}$
            \EndIf
        \EndFor
    \EndFor
\EndFor

\ForAll{$e \in \Gamma$}
    \ForAll{$\pi \in \mathcal{C}$}
        \If{$\tau \in \pi$}
            \If{$e$ has $\nu$}
                \State $\mathcal{R} \gets \mathcal{R} \cup \{(d, \tau, \pi_{\text{id}}, \nu)\}$
                \Comment{$\pi_{\text{id}}$: Pathway ID}
            \Else
                \State $\mathcal{R} \gets \mathcal{R} \cup \{(d, \tau, \pi_{\text{id}})\}$
            \EndIf
        \EndIf
    \EndFor
\EndFor

\State \textbf{return} $\mathcal{R}$
\end{algorithmic}
\end{algorithm}
\begin{figure*}[h]
  \centering
  \includegraphics[width=\linewidth]{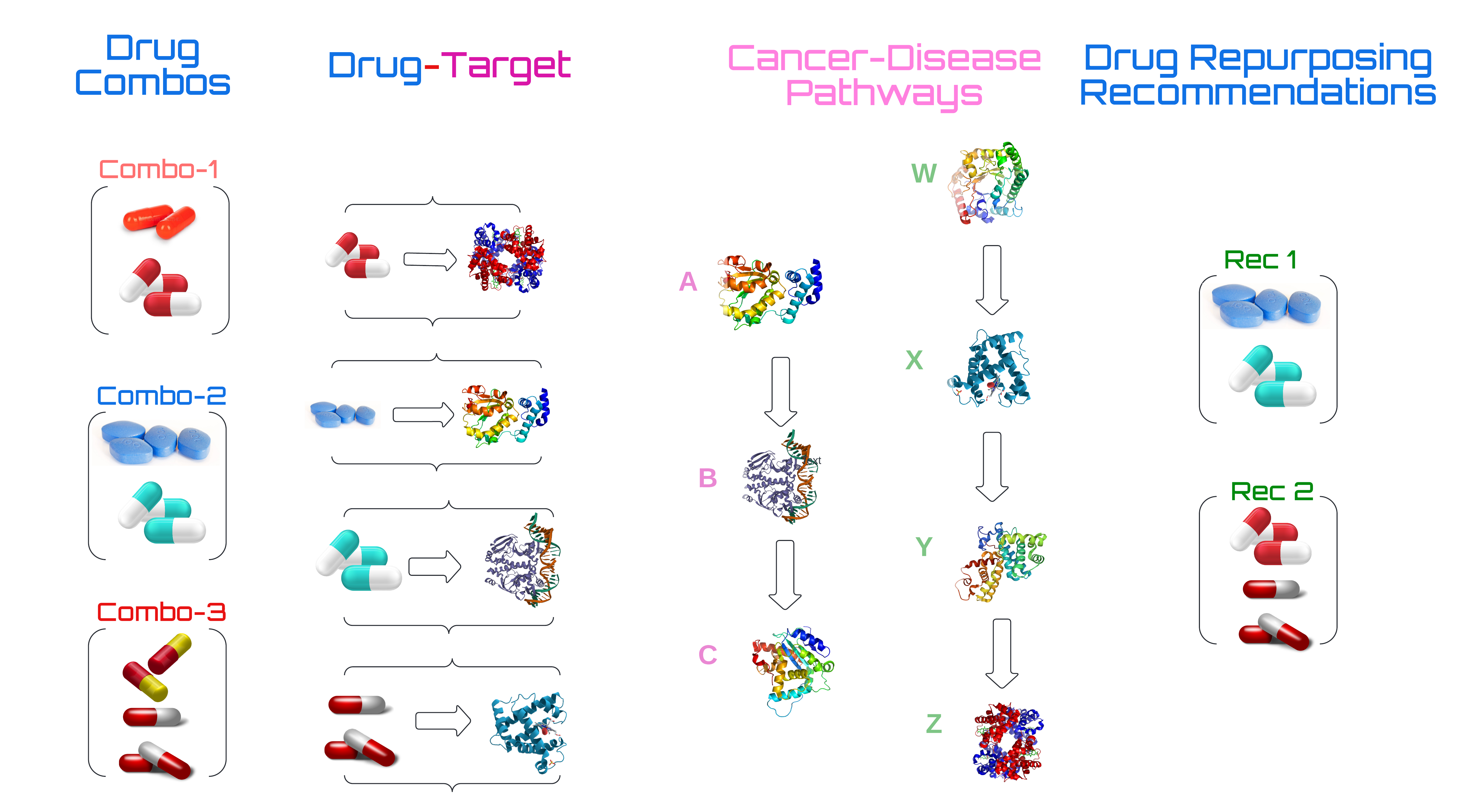}
  \caption{The Algorithmic Process of Discovering Repurposable Drug Combinations}\label{fig:combo-disco}
\end{figure*}

\section{Results}

\subsection{The Prompt-Engineering Outcome}
The few-shot prompt engineering against 2496 clinical trials resulted in 3281 drugs and 4502  ``co-occurrence'' relationship among each pair. We found that 2754 of the co-occurrence relationships were semantically encoded as ``combination therapy'', which signaled that the drugs that make up the two ends of the link are suggested as a combination. Since we aim to identify combinations from FDA-approved drugs, we validated drug members of the combinations using the OpenFDA database. The outcome of such validation confirmed that 2680 nodes identified by ChatGPT as recognizable, while the remaining 601 nodes were terms that are either not FDA-approved drugs or simply noise. 

From the terms identified by ChatGPT as FDA-approved and recognizable (e.g., 'Oral anti-diabetics', 'Antiretroviral Therapy', 'Psychotherapy', 'Chemotherapy treatment', 'HIV medications'), it's clear these are descriptive terms for drugs, but not specific for considering drug combinations. However, a small number of identified drugs, despite being part of FDA-approved terms, turned out to be false positives. ChatGPT mistakenly identified terms such as ('HIV infection', 'Merkel cell cancer', 'Prostate Cancer') as drugs.

Conversely, many terms identified by ChatGPT that were not recognized by OpenFDA were indeed drugs. This discrepancy arises because OpenFDA lists only drugs that are already approved. Drugs identified by ChatGPT that follow a pattern like manufacturing initials followed by digits (e.g., AZD6738 by AstraZeneca, BMS-791325 by Bristol-Myers Squibb, JNJ-42847922 by Johnson and Johnson, TAK-491 by Takeda) are typically still under investigation and have not yet been approved for production.

Since the outcome of the prompt-engineering step forms a foundational block for subsequent steps, our focus is on removing false positives and confirming true negatives. The consideration of drugs under investigation requires careful inclusion. Table \ref{tab:fda-validation} show the general outcome of the FDA-approved. 

\begin{table}[ht]
\caption{The description of drug combinations graph and the outcome of the validation process using the OpenFDA database.}
\label{tab:fda-validation}
\centering
\begin{tabular}{cccc} % Four columns, centered
    \toprule
    \textbf{\#Nodes} & \textbf{\# Edges } & \textbf{\#Validated Nodes} & \textbf{\# Validated Edges} \\
    \midrule
    3281 & 4502 & 2680 & 2754 \\
    \bottomrule
\end{tabular}
\end{table}

\subsection{The Outcome of Generating of Drug-Target Combinations Algorithm}
The discovery process executed by Algorithm \ref{alg:drug_combinations_graph}yielded 18,293 unique PubMed documents, covering drug-target relationships across 36-46 breast cancer signaling pathways. This coverage was explored using 5 different proximity values (10, 20, 30, 40, and 50 tokens) as they appeared in the original PubMed abstracts. The process captured each signaling pathway along with the proteins comprising it, drugs associated with those proteins as potential targets, and the proximity between each drug and target in the corresponding PubMed abstracts.

A common pattern observed was that shorter proximity values correlated with fewer occurrences between a drug and a protein within a given signaling pathway, and fewer supporting PubMed abstracts for such connections. Contrary to expectation, not all cases showed an increase in the number of proteins, as seen in pathway ID hsa:2353.

Table \ref{tab:proximity-analysis} presents 20 signaling pathways grouped by proximity. Each group includes the number of supporting PubMed abstracts, the count of drugs, and the mentions of proteins as potential targets. The table highlights pathways such as signaling pathway ID hsa:10000, supported by numerous drugs but covering only one protein.

\begin{figure*}[htbp]
    \centering
    \begin{subfigure}[b]{0.48\textwidth}
        \includegraphics[width=\textwidth]{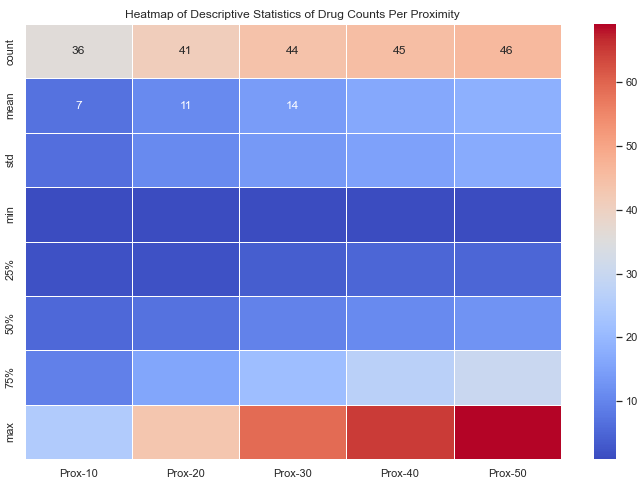}
        \caption{Drug Heatmap}
        \label{fig:drug_heatmap}
    \end{subfigure}
    \hfill
    \begin{subfigure}[b]{0.48\textwidth}
        \includegraphics[width=\textwidth]{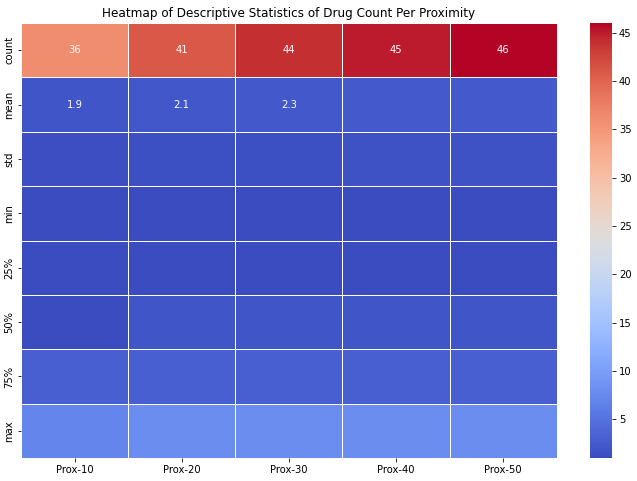}
        \caption{Protein Heatmap}
        \label{fig:protein_heatmap}
    \end{subfigure}
    \caption{Drug and Protein Heatmaps}
    \label{fig:drug_protein_heatmaps}
\end{figure*}

\begin{table*}[htbp]
\centering
\captionsetup{justification=centering}
\caption{An overview of the frequency of PubMed, drug, and target occurrences in drug-target associations is provided}
\label{tab:proximity-analysis}
\begin{tabular}{@{}l||lll|lll|ccc@{}}
\toprule
\multirow{2}{*}{Pathway} & \multicolumn{3}{c|}{Prox-10} & \multicolumn{3}{c|}{Prox-30} & \multicolumn{3}{c}{Prox-50} \\
& \multicolumn{1}{c}{No. PubMed} & \multicolumn{1}{c}{No. Drugs} & \multicolumn{1}{c|}{No. Proteins} & \multicolumn{1}{c}{No. PubMed} & \multicolumn{1}{c}{No. Drugs} & \multicolumn{1}{c|}{No. Proteins} & \multicolumn{1}{c}{No. PubMed} & \multicolumn{1}{c}{No. Drugs} & \multicolumn{1}{c}{No. Proteins} \\ \midrule \midrule
hsa:2099 & 33 & 25 & 4 & 79 & 59 & 4 & 98 & 69 & 4 \\
hsa:1956 & 27 & 20 & 4 & 49 & 37 & 5 & 62 & 42 & 6 \\
hsa:8202 & 1 & 1 & 1 & 9 & 6 & 4 & 17 & 11 & 6 \\
hsa:5604 & 5 & 5 & 2 & 6 & 6 & 2 & 9 & 9 & 2 \\
hsa:5594 & 10 & 9 & 4 & 37 & 26 & 5 & 49 & 35 & 6 \\
hsa:182 & 2 & 2 & 1 & 6 & 8 & 2 & 9 & 12 & 2 \\
hsa:4851 & 3 & 3 & 1 & 5 & 5 & 1 & 8 & 8 & 1 \\
hsa:2475 & 17 & 17 & 1 & 25 & 24 & 2 & 27 & 31 & 2 \\
hsa:3479 & 10 & 5 & 3 & 16 & 11 & 3 & 24 & 16 & 3 \\
hsa:3480 & 1 & 1 & 1 & 6 & 5 & 2 & 10 & 8 & 2 \\ 
hsa:2064 & 48 & 22 & 7 & 86 & 39 & 8 & 113 & 51 & 8 \\
hsa:2353 & 8 & 7 & 3 & 19 & 14 & 3 & 24 & 15 & 3 \\
hsa:5241 & 14 & 12 & 2 & 57 & 42 & 2 & 66 & 50 & 2 \\
hsa:5290 & 16 & 11 & 3 & 35 & 28 & 4 & 47 & 33 & 4 \\
hsa:10000 & 1 & 1 & 1 & 2 & 2 & 1 & 4 & 5 & 1 \\
hsa:5728 & 6 & 6 & 3 & 17 & 16 & 3 & 30 & 28 & 3 \\
hsa:595 & 2 & 2 & 1 & 9 & 10 & 1 & 14 & 14 & 2 \\
hsa:4609 & 11 & 10 & 2 & 29 & 23 & 2 & 42 & 34 & 2 \\
hsa:1019 & 3 & 3 & 1 & 8 & 8 & 1 & 12 & 13 & 1 \\
hsa:5925 & 12 & 9 & 3 & 28 & 22 & 3 & 34 & 26 & 3 \\ \bottomrule
\end{tabular}
\end{table*}
Taking a closer look to the summary of the statistics of the covered drugs and their potential target. On the one hand, the analysis shows that the means of the drug frequencies increased as the distance also increased (7, 11, 14, 17, 18), while the max number of drug found, also exhibited the same behavior, ranging from (25, 43,  59,  65,  69). On the other hand, the means of protein targets frequencies were much lower in value than drug, which ranged in values (1.9, 2.1,  2.29, 2.4, 2.4) for each of the proximity respectively. While the max frequencies were much lower than the counterparts of drugs (7, 8, 8, 8, 8). Overall, the ratios between drug frequencies vs protein frequencies in terms of mean and max are (0.27, 0.19, 0.16, 0.14, 0.13) and (0.28, 0.19, 0.14, 0.12, 0.12) respectively. Figure \ref{fig:drug_protein_heatmaps} is a heatmaps that shows the summary of this drug protein frequencies for each of the five proximity parameters. 
As for the PubMed supporting evidence for the drug-target pairs, we found that the means of supporting number of documents are (9, 14, 17, 20, 23) while the max (48, 72, 86, 98, 113). Here, we observe that the Frequencies of PubMed supporting documents increase as the proximity distance also increases. By measuring the mean-to-maximum ratio which produces a ratio of (0.18, 0.19, 0.19, 0.20, 0.20) which indicate that means increased higher that the max as the distance increased. This is explained the PubMed dataset has a finite number of documents that support drug-targets at some point reaches its maximum and the distance becomes irrelevant. 
\begin{figure}[htbp]
    \centering
    \includegraphics[width=0.5\textwidth]{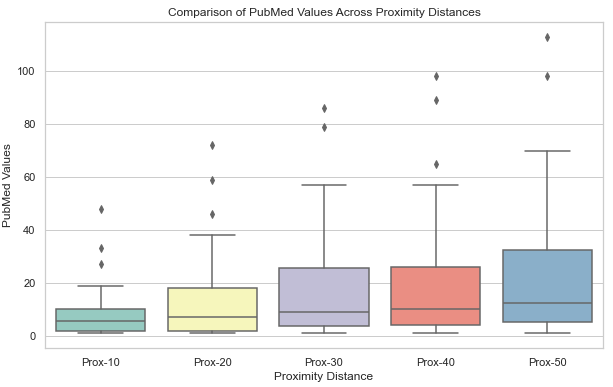}
    \caption{Comparison of PubMed Frequencies Across Proximity Distances}
    \label{fig:violin_plot}
\end{figure}
The drug combination layer that was previously produced by the ChatGPT prompt-engineering process produced a graph of 3281 nodes and 4502 edges. The breast cancer pathways is comprised of 46 pathways which were supported by 1178 drugs where at least 2 drugs supported each pathway. The execution of Algorithm \ref{fig:combo-disco} resulting in confirming that 980 drugs discovered from the literature had a target on the breast cancer 46 pathways, while 198 drugs had no association with any protein targets of any of the pathways. Table \ref{table:evidence} captures the summary of each pathway and how many drugs are supporting each path. While hsa:2064 and hsa:2099 have the most coverage from drugs (108, and 91), pathways such as hsa:6654 and hsa:1499 has the fewest number of drugs as coverage from drugs (3 and 2). It is our intuition such the prediction of drugs combinations for the pathways that has low coverage would be most valuable as it seems that little is known about such pathways. 
\begin{table}[h!]
\centering
\caption{Path ID and Drug Count Coverage}
\label{table:evidence}
\begin{tabular}{| l | r | l | r |}
  \hline
  \textbf{Path ID} & \textbf{Drug Count} & \textbf{Path ID} & \textbf{Drug Count} \\
  \hline
  \hline
  hsa:2099  & 91  & hsa:4609  & 37  \\
  hsa:1956  & 68  & hsa:1499  & 2   \\
  hsa:8202  & 10  & hsa:1019  & 10  \\
  hsa:3265  & 3   & hsa:5925  & 32  \\
  hsa:5604  & 10  & hsa:1869  & 8   \\
  hsa:5594  & 43  & hsa:3815  & 38  \\
  hsa:2885  & 5   & hsa:1026  & 42  \\
  hsa:6654  & 3   & hsa:2260  & 4   \\
  hsa:182   & 11  & hsa:8600  & 9   \\
  hsa:4851  & 5   & hsa:1950  & 16  \\
  hsa:2475  & 29  & hsa:672   & 34  \\
  hsa:6198  & 4   & hsa:675   & 26  \\
  hsa:3479  & 25  & hsa:2246  & 3   \\
  hsa:3480  & 7   & hsa:2324  & 14  \\
  hsa:2064  & 108 & hsa:7157  & 64  \\
  hsa:2353  & 19  & hsa:581   & 26  \\
  hsa:5241  & 66  & hsa:578   & 13  \\
  hsa:5290  & 45  & hsa:10000 & 4   \\
  hsa:5728  & 28  & hsa:595   & 12  \\
  \hline
  \hline
\end{tabular}
\end{table}

\section{Discussion}
This paper introduces a novel Generative AI approach to prompt engineering, leveraging ChatGPT with a few-shot training set. While the final results demonstrated high accuracy, achieving satisfactory prompts required multiple iterations. The primary goal was to extract drug names and identify their potential combinations as mentioned in clinical trials. The prompt was dynamically constructed using configuration parameters: (1) the actual clinical trial descriptions, (2) examples teaching ChatGPT drug names and how combinations are indicated (e.g., by symbols like "+"), and (3) instructions for ChatGPT to format responses in a specific manner — specifically, pairing NCT\_ID with two drugs per line.

Despite learning from the provided examples, ChatGPT occasionally deviated from instructions by listing more than two drugs on a line. We also experimented with structured formats like XML and JSON, but encountered challenges with producing valid outputs consistently. Interestingly, ChatGPT successfully recognized drug names and categories, even when they were not explicitly concrete drug names.

In extracting drug-target links from biomedical abstracts, each pair was encoded with proximity distance measured in tokens. Proximity proved effective in filtering noise by reducing unnecessary candidates while enhancing coverage when information about signaling pathways, proteins, and drug targets was limited. The adjustability of proximity is crucial, as closer distances typically indicate higher relevance.

The study focused on utilizing KEGG breast cancer signaling pathways as a case study for our framework to recommend drug opportunities. However, while some signaling pathways were well-covered with over 100 candidates, others had minimal coverage, with as few as one candidate. These were not included in Table \ref{table:evidence} as our framework specializes in generating drug combinations. It is important to further investigate the significance of these proteins as potential targets and their roles in cancer progression. Future enhancements may involve integrating data from other databases such as the STRING database \cite{szklarczyk2021string} and the IntAct database \cite{hermjakob2004intact} to explore new protein-protein interactions targeted by existing drugs, potentially offering novel treatment strategies.

\section{Conclusion and Future Directions}
In this paper, we introduced a multilayered network medicine framework empowered by Generative AI to expedite drug development and discovery. Our framework specializes in predicting potentially repurposable drug combinations for complex diseases in general, and we demonstrated a case study for breast cancer therapy. Utilizing literature-based drug-target co-occurrences, we algorithmically explored and identified drugs targeting multiple proteins across the 46 KEGG signaling pathways associated with breast cancer, indicating potential combinations.

This research pushed the boundaries of Generative AI by leveraging ChatGPT for prompt engineering against real-world evidence sourced from clinical trials. We developed a parameterized system capable of dynamically analyzing clinical trial data to generate prompt instances, enriched with learning from a few-shot training approach. While ChatGPT successfully recognized all drugs mentioned in clinical trials and literature, it faced challenges in accurately identifying drug combinations as instructed by the prompt. We mitigated this issue by programmatically constructing the combinations.

Looking forward, our framework aims to expand its scope to investigate drug repurposing across various cancer types. Initially, we plan to focus on prostate cancer and lung cancer, integrating relevant literature and KEGG signaling pathways specific to these diseases. Additionally, ongoing efforts will involve refining ChatGPT prompts to enhance prediction accuracy of drug combination relationships from diverse sources.

To ensure safe and effective combination therapies with minimal adverse effects, we intend to conduct further validation of drug similarities. This process will prevent overly similar drugs from being combined, thereby reducing potential toxicity and side effects.

Ultimately, the final product of this research will be the deployment of our findings in a ChatGPT-based interactive service. This service will enable users to inquire about predicted drugs for repurposing, signaling pathway support for target proteins, and other relevant data pertaining to drug combinations. To ensure traceability, explainability, and trust, to follow some of most informing policies \cite{hamed2024safeguarding}, on how combinations were determined, we document the source of each combination from clinical trials and PubMed abstracts, specifying the proteins within specific signaling pathways that the drugs are likely to bind with.
\section{Acknowledgments}
This research is not yet funded and this work is done as foundational for future funding. The authors would like to thank Dr. Pawel Mikulski of iMol PAS and Dr. Sachin Kote of Gdansk University for the valuable discussions.
\bibliographystyle{unsrt}
\bibliography{references}
\end{document}